\documentclass{article}

\PassOptionsToPackage{numbers}{natbib}

\usepackage[final]{neurips_2023}

\usepackage[utf8]{inputenc} 
\usepackage[T1]{fontenc}    
\usepackage{hyperref}       
\usepackage{url}            
\usepackage{booktabs}       
\usepackage{amsfonts}       
\usepackage{nicefrac}       
\usepackage{microtype}      
\usepackage[dvipsnames]{xcolor}         
\usepackage[inline]{enumitem}
\usepackage{amsmath}
\usepackage{todonotes}
\usepackage{caption}
\usepackage{subcaption}
\usepackage{multirow}
\usepackage{wrapfig}
\usepackage{todonotes}

\title{Efficient infusion of self-supervised representations in Automatic Speech Recognition}

\author{%
  Darshan Prabhu\thanks{Work done during an internship at Sony Research India. Corresponding to: Sai Ganesh Mirishkar} \\ Sony Research India\\ darshan.prabhu@sony.com  \And {} Sai Ganesh Mirishkar \\Sony Research India \\ saiganesh1.mirishkar@sony.com  \And {} Pankaj Wasnik \\ Sony Research India \\ pankaj.wasnik@sony.com 
}

\begin{document}

\maketitle

\begin{abstract}
Self-supervised learned~(SSL) models such as Wav2vec and HuBERT yield state-of-the-art results on speech-related tasks. Given the effectiveness of such models, it is advantageous to use them in conventional ASR systems. While some approaches suggest incorporating these models as a trainable encoder or a learnable frontend, training such systems is extremely slow and requires a lot of computation cycles. In this work, we propose two simple approaches that use 
\begin{enumerate*}[label={(\arabic*)}]
    \item framewise addition and
    \item cross-attention
\end{enumerate*}
 mechanisms to efficiently incorporate the representations from the SSL model(s) into the ASR architecture, resulting in models that are comparable in size with standard encoder-decoder conformer systems while also avoiding the usage of SSL models during training. Our approach results in faster training and yields significant performance gains on the Librispeech and Tedlium datasets compared to baselines. We further provide detailed analysis and ablation studies that demonstrate the effectiveness of our approach.


\end{abstract}

\section{Introduction} 

Since speech is a complex signal, audio-related tasks like Automatic Speech Recognition(ASR) rely heavily on having robust speech representations as input. While standard ASR systems use conventional signal processing techniques~\cite{ft} for generating these representations, another set of models, called representation models, are tasked with learning to generate such representations using large unlabeled data. These models are trained using the masked language modeling~\cite{devlin2019bert} objective, where the model attempts to reconstruct parts of the text/audio that are masked. As a result, they can understand raw text/speech well and generate high-quality text/speech representations. BERT~\cite{devlin2019bert}, Wav2Vec~\cite{baevski2020wav2vec}, and HuBERT~\cite{hsu2021hubert} are some popular models that fall under this category. They are also called Self-Supervised Learned(SSL) models, as they do not need labeled data during training. These models are popular for downstream tasks as they perform exceptionally well even with small amounts of data. However, since these models are huge, training an ASR system retrofitted with such models as a learnable frontend~\cite{zeghidour2021leaf} or an Encoder with limited computational resources becomes challenging, if not impossible. To address this issue, several works propose freezing the representation model and using its output as auxiliary information to a custom encoder. This enables us to entirely forgo the usage of SSL models during training, as the extraction of these representations can be offloaded to preprocessing~\footnote{This is a one-off step performed once at the beginning of training.}. Early work in this regard is found in Natural Language Processing(NLP), where BERT representations are integrated into the Neural Machine Translation system~\cite{zhu2020incorporating}. Recent works perform this integration for multimodal inputs as well~\cite{zhao2022multilevel,KIM2022168}. In speech, fine-tuned Wav2vec2 embeddings have been used as auxiliary data alongside the domain adversarial setup~\cite{li2021accentrobust,mirishkar2023iiith} for the accented ASR setting.

In this work, we propose two approaches that incorporate representations from pre-trained SSL models into an end-to-end ASR architecture. We also explore the possibility of introducing multiple such representations into the architecture and perform a comprehensive analysis of our approach in terms of convergence, efficiency, and understanding.

\section{Proposed Methodology}

\begin{figure}[t]
\centering
\hfill
\begin{subfigure}{\textwidth}
  \centering
  \includegraphics[width=\linewidth]{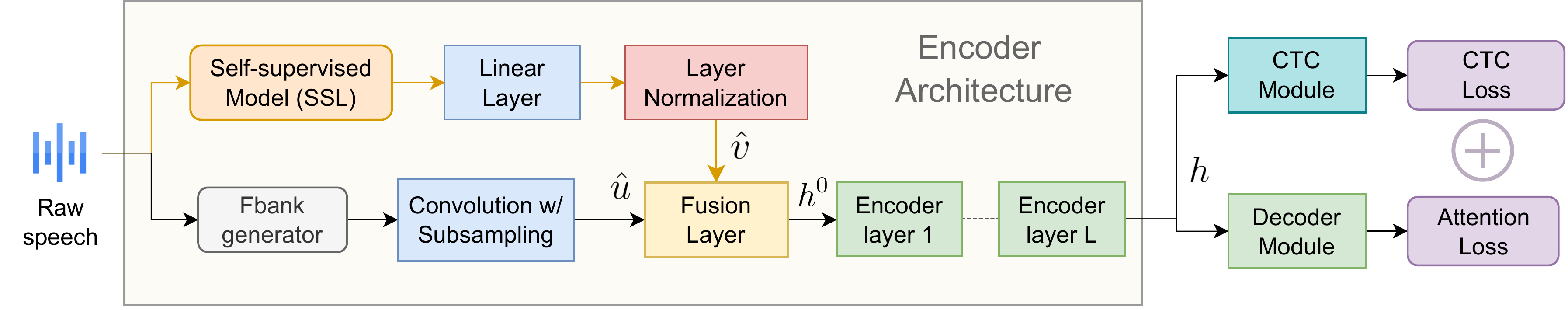}
  \caption{Overall Architecture}
  \label{fig:overall}
\end{subfigure}%
\begin{subfigure}{.01\textwidth}
\end{subfigure}
\begin{subfigure}{.4\textwidth}
  \centering
  \includegraphics[width=.7\linewidth]{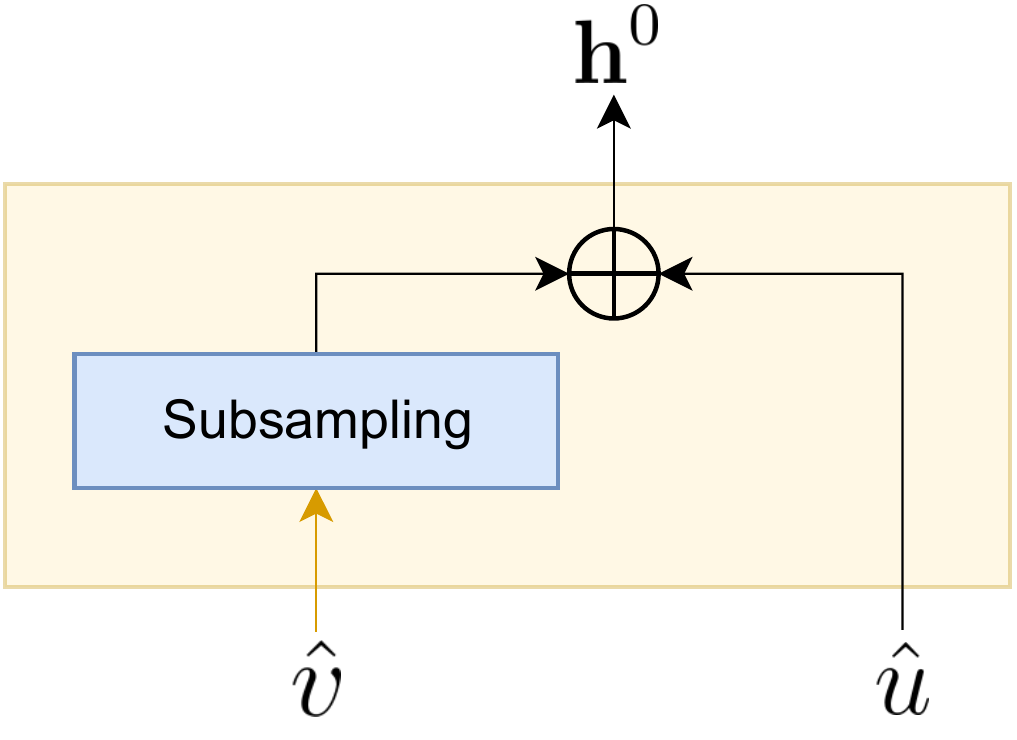} \\
  \caption{Subsampled Framewise Addition (SFA)}
  \label{fig:sfa}
\end{subfigure}
\begin{subfigure}{.4\textwidth}
  \centering
  \includegraphics[width=.7\linewidth]{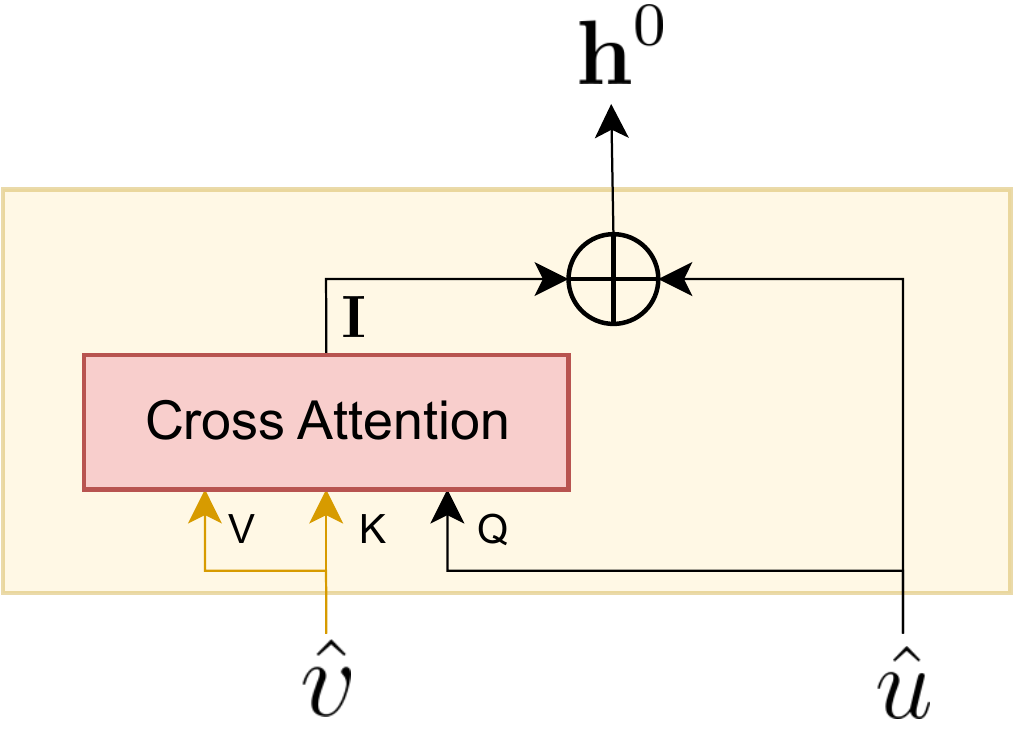}
  \caption{Cross Attention (CA)}
  \label{fig:ca}
\end{subfigure}
\hfill
\caption{Overview of our proposed Architecture that integrates the representations from the self-supervised model into the ASR encoder using two approaches: Subsampled Framewise Addition(\textbf{SFA}) and Cross Attention(\textbf{CA}). {\large $\hat{v}$} and {\large $\hat{u}$} are the self-supervised and fbank representations respectively. }
\label{fig:architecture}
\end{figure}

Figure~\ref{fig:architecture} shows our proposed modifications to the existing joint CTC-Attention~\cite{kim2017joint} framework that allows auxiliary audio representations(extracted from pre-trained SSL models) to be easily introduced within the end-to-end ASR architecture without substantially increasing the model size~\footnote{As the usage of SSL model during inference is unavoidable, no computational benefit would be observed using our approach for inference. }. Figure~\ref{fig:overall}  shows the overall architecture that comprises three main modules: Encoder(\textsc{Enc}), Decoder(\textsc{Dec-Att}) and Connectionist Temporal Classification(\textsc{Dec-Ctc})~\cite{ctc}. Although all our changes are made to the encoder, we first briefly explain the overall architecture, followed by our proposed modifications to the encoder.

Let $\textbf{x}=\{ x_1, x_2, \ldots, x_N \}$ be the raw representation of the input audio. This representation $x$ is passed through 
\begin{enumerate*}[label={(\arabic*)}]
    \item Fbank generator that generates $T$ length fbank representation $\textbf{u}=\{ u_1, u_2, \ldots, u_T \}$ where $u_i \in \mathbb{R}^d$ and
    \item Self supervised pre-trained model that generates $T'$ length SSL representation $\textbf{v}=\{ v_1, v_2, \ldots, v_{T'} \}$ where $v_i \in \mathbb{R}^{d'}$
\end{enumerate*}. 
The encoder(\textsc{Enc}) jointly reasons on both $\textbf{u}$ and $\textbf{v}$ to generate contextualized audio representation $\textbf{h} = \textsc{Enc}(\textbf{u},\textbf{v}) = \{ h_1, h_2, \ldots h_T \}$. This contextualized representation $\textbf{h}$ is then consumed by \textsc{Dec-Att} and \textsc{Dec-Ctc} module which aim at predicting the output token sequence $\textbf{y} = \{ y_1, y_2, \ldots, y_i, \ldots y_M \}$ using CTC and Attention loss. 


The encoder module begins with a $\mathrm{Convolution\_Subsampling}$ block that applies convolution and two-factor subsampling~\footnote{Subsampling is a crucial step that reduces the computation complexity of the encoder while performing at par with the original encoder~\cite{burchi2021efficient}.} on the fbank representation $\textbf{u}$ resulting in a $T/2$ length sequence $\hat{\textbf{u}}$. Parallely, we employ a linear layer and layer normalization on the SSL representation $\textbf{v}$ to obtain a $T'$ length sequence $\hat{\textbf{v}}$. While layer normalization helps with better generalization, the linear layer reduces the dimension from $d'$ to $d$.
These sequences are then fed to the $\mathrm{Fusion\_Layer}$ that acts on both these inputs, performing a deterministic fusion of $\hat{\textbf{u}}$ and $\hat{\textbf{v}}$ to generate $\textbf{h}^0$(In section \ref{arch:fusion}, we discuss in detail the different choices available for the fusion layer along with their merits and demerits.).
The output of the fusion layer is then passed through a stack of $L$ identical encoder layers. Each encoder layer feeds on the output from the previous layer $\textbf{h}^{i-1}$ and generates contextualized representation $\textbf{h}^i$ by performing the standard operation of a conformer~\cite{gulati2020conformer} encoder layer.

\subsection{Fusion Layer}\label{arch:fusion}
In this section, we discuss two simple approaches to generate the representation $\textbf{h}^0$ using sequences $\hat{\textbf{u}}$ and $\hat{\textbf{v}}$. To reiterate, $\textbf{u}$ is a $T$ length sequence that is passed through two-factor subsampling to generate the representation $\hat{\textbf{u}}$ which is a sequence of length $T/2$. $\hat{\textbf{v}}$ which is obtained by passing raw audio through the SSL model followed by layer normalization is a $T'$ length sequence which in the case of Wav2Vec and HuBERT happens to be equal to $T$. Finally, the output of this fusion layer is $\textbf{h}^0$ which is also a $T/2$ length sequence.

\paragraph{Subsampled Framewise Addition (SFA):} Inspired by the work of Jialu et al.~\cite{li2021accentrobust}, we first propose a simple parameterless approach as shown in Figure~\ref{fig:sfa} that relies on the observation that performing subsampling on $\hat{\textbf{v}}$ by a factor of 2 conveniently leads to both $\hat{\textbf{u}}$ and $\hat{\textbf{v}}$ to be of equal length. We can then perform framewise addition of both these sequences to generate $\textbf{h}^0$. Mathematically, this generation can be written as: $ \textbf{h}^0_i = \hat{\textbf{u}}_i \oplus \hat{\textbf{v}}_{min(T, 2 \times i )}(\forall \enspace 1 \le i \le T/2)$ 
where $\textbf{h}^0_i$ is the $i^{th}$ entry in $\textbf{h}^0$ and $\oplus$ is the elementwise addition of two $d$-dimensional vectors. As this is a parameterless operation, it does not add to the model size but there are two main drawbacks with this approach. First, we have a predetermined decision on which frame of the corresponding sequences are to be added. It would be beneficial to let the model determine this mapping. Second, the approach heavily relies on the lengths of the two sequences to have some linear relation, and for arbitrary lengths, it becomes harder to determine this mapping. It becomes exponentially harder if the lengths of both sequences differ by a large margin. 

\paragraph{Cross Attention (CA):} To address both the concerns, we introduce a cross-attention layer as shown in Figure~\ref{fig:ca} that uses the concept of attention to determine how each frame of $\hat{\textbf{u}}$ wants to attend to the frames of $\hat{\textbf{v}}$. Mathematically, this generation of $\textbf{h}^0$ can be written as:
\begin{center}
    \begin{math}
        \textbf{h}^0_i = \hat{\textbf{u}}_i \oplus \textbf{I}_i \quad \forall \quad 1 \leq i \leq T/2 \enspace \textrm{where} \enspace \textbf{I} = \mathrm{MultiHeadAttention}(\hat{\textbf{u}}, \hat{\textbf{v}}, \hat{\textbf{v}})
    \end{math}
\end{center}
where $\mathrm{MultiHeadAttention}(Q, K, V)$ refers to the standard multi-headed attention proposed by Vaswani et al.~\cite{vaswani2023attention} with $Q, K$ and $V$ denoting query, key and value respectively. This approach is free from any predetermined mapping of the frames of the corresponding sequences, as attention helps the model learn such mappings.

\section{Experimental Setup}

We run all our experiments on NVIDIA A100 GPUs using the ESPnet toolkit~\cite{espnet}. As is common practice, we add 3-way speed perturbation to both datasets before training. Unless specified otherwise, in all of our experiments, we train a conformer model with 12 encoder and 6 decoder layers. We use 256-dimensional tensors and four heads for attention computation. All the models are trained for 50 epochs with a batch size of 32, a learning rate of 1.0, and 4 gradient accumulation steps. The representations from SSL models are dumped beforehand to speed up the training. Throughout all of our experiments, we use three self-supervised models: (1) Wav2vec 2.0 \textsc{Base}, (2) HuBERT \textsc{Base}, and (3) HuBERT \textsc{Large}. All these models are only pre-trained with the SSL objective without any ASR fine-tuning. We report numbers on Librispeech-100~\cite{librispeech} and Tedlium2~\cite{tedlium} datasets that are publicly available.

\section{Results}

\setlength{\tabcolsep}{3pt}
\begin{table}[t]
  \caption{Comparison of Performance(WER) of our system with baselines on Librispeech-100 dataset~\cite{librispeech}. Our experiments are labeled in the format $X$+$Y$+$Z$, indicating that the architecture $X$ was modified to include representation from the $Y$ SSL model by employing the $Z$ approach.}
  \vspace{0.2cm}
\label{res:librispeech}
\centering
  \begin{tabular}{lcccc}
        \toprule
        Method & \small{Dev Clean} & \small{Dev Other} & \small{Test Clean} & \small{Test Other} \\
        \midrule

        \small{B1: Transformer (Trans.)} & 10.1 & 25.8 & 10.4 & 26.4 \\[1pt]
        \small{B2: Conformer (Conf.)} & 7.9 & 21.4 & 8.4 & 22.0 \\[1pt]
        \midrule
        \small{T1: Conf. + w2v-\textsc{Base} + SFA} & 5.9 & 16.2 & 6.4 & 16.4 \\[1pt]
        \small{T2: Conf. + w2v-\textsc{Base} + CA} & 5.9 & 16.0 & 6.3 & 16.3 \\[1pt]
        \small{T3: Conf. + HuBERT-\textsc{Base} + SFA} & 5.2 & 13.5 & \textbf{5.4} & 13.5 \\[1pt]
        \small{T4: Conf. + HuBERT-\textsc{Base} + CA} & \textbf{\small{5.1}} & \textbf{13.0} & \textbf{5.4} & \textbf{13.3} \\[1pt]
        \small{T5: Conf. + (w2v-\textsc{Base}, HuBERT-\textsc{Base}) + CA} & 5.7 & 14.1 & 6.1 & 14.5 \\[1pt]
        \bottomrule
  \end{tabular}
\end{table}

Table~\ref{res:librispeech} compares word error rates (WER) between our proposed architectures (T1 to T5) and baselines (B1 and B2) on the Librispeech 100h~\cite{librispeech} dataset. We report two baselines, out of which the Conformer baseline outperforms the Transformer baseline by a large margin. We show five variants of our architecture, labeled T1 through T5, each of which uses the conformer architecture but differs in either the choice of SSL model or the fusion method. We see that among Wav2vec and HuBERT, HuBERT representations consistently perform better. Furthermore, we find no added advantage in using both Wav2Vec and HuBERT representations. Irrespective of the SSL model chosen, cross-attention(CA) shows significant improvement over the sub-sampled framewise addition(SFA) approach.

\section{Analysis}

\begin{figure}[h]
\vspace{0.03cm}
\begin{minipage}[c]{0.42\linewidth}
\includegraphics[width=\linewidth]{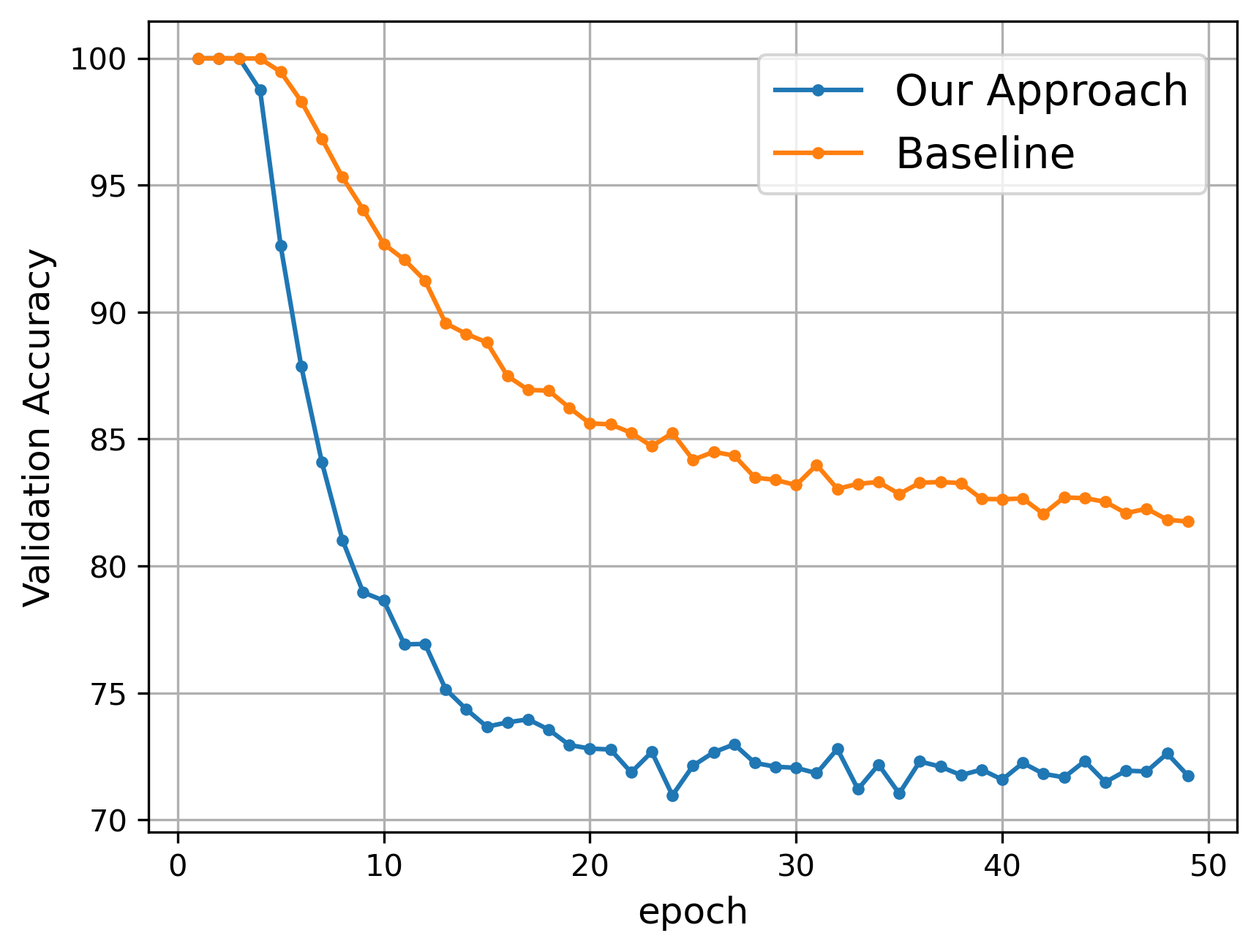}
\caption{Comparison of epochwise model performance(WER) between baseline and our best setting({\scriptsize Conf.+HuBERT-\textsc{Base}+CA}) on the validation split of Librispeech-100 dataset.}
\label{analysis:wer}
\end{minipage}
\hfill
\begin{minipage}[c]{0.55\linewidth}
\includegraphics[width=\linewidth]{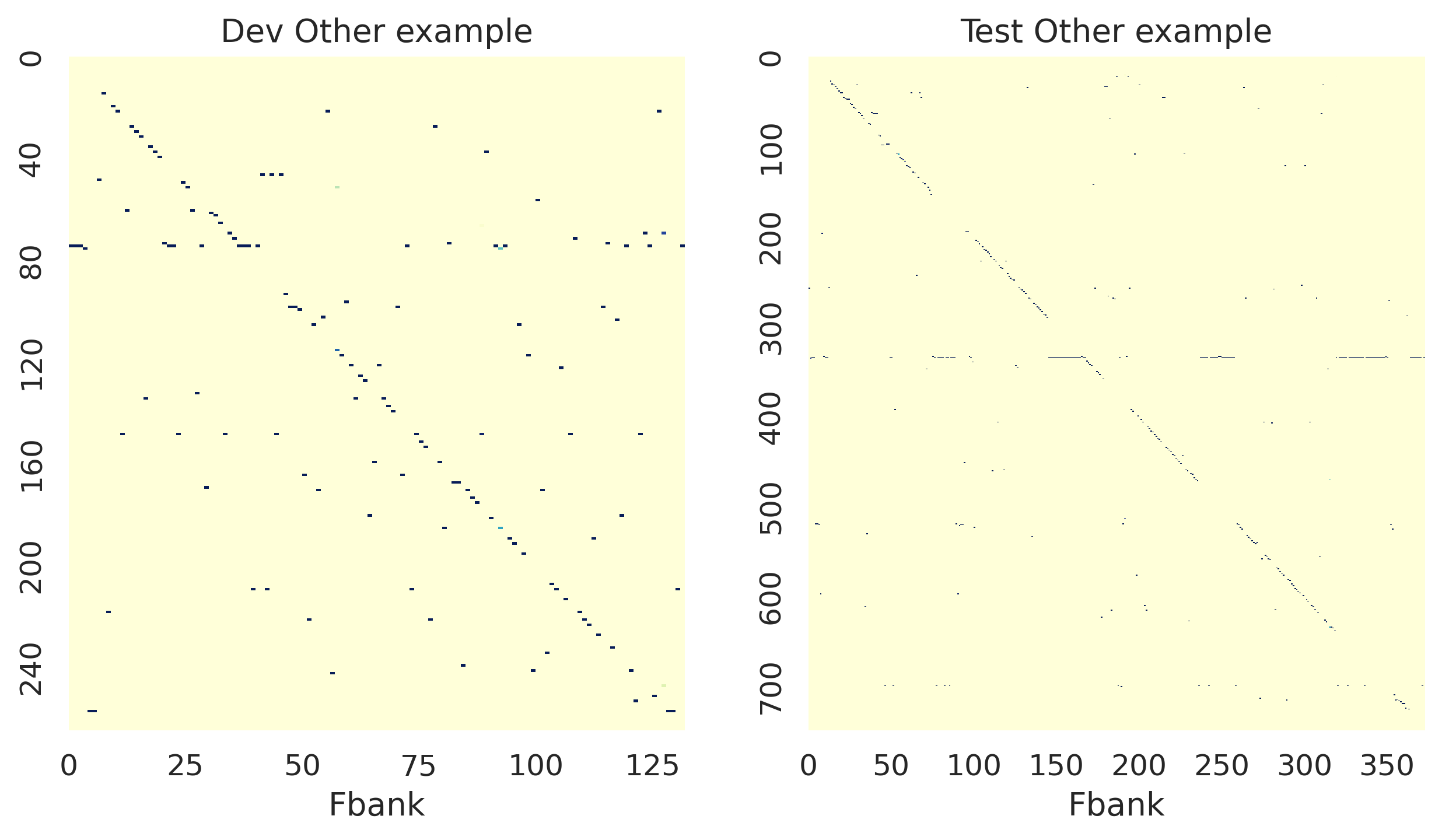}
\caption{Visualization of attention scores for two samples from the Librispeech-100 dataset, one from Dev-Other split and Test-Other split respectively. We use the model from our best setting to obtain these scores.}
\label{analysis:heatmap}
\end{minipage}%
\end{figure}

\paragraph{Faster Model Convergence:} 
As shown in Figure~\ref{analysis:wer}, our approach outperforms the baseline best results within only ten epochs of training and continues to improve over the next few iterations. Moreover, the convergence is much faster in the early iterations, resulting in a reasonably good model within the first 5-10 epochs of training. 

\paragraph{Efficacy of our approach:} 
While directly incorporating HuBERT or Wav2vec as a frontend/Encoder is bound to perform better, our approach results in a model that is both faster in training and smaller in size at the cost of a slight degradation in performance. Furthermore, performing SSL feature extraction in the preprocessing stage adds a very minimal overhead~\footnote{This process can be easily parallelized. In our experiments, we use sixty parallel jobs to perform feature extraction.} ( approx 4 hrs for Librispeech-100 and 10 hrs for Tedlium2 ) compared to using SSL models throughout training.  

\paragraph{Number of encoder layers:} Table~\ref{res:analysis_encoder} compares the conformer baseline with four variations of our best system({\small Conf.+HuBERT-\textsc{Base}+CA}) which differ only in the number of encoder layers used. Even after reducing the number of encoder layers by 80\%, we find that the model still performs much better despite having only half as many parameters and training time as the baseline model. 

\setlength{\tabcolsep}{1.3pt}
\begin{table}[h]
    \begin{minipage}{0.64\textwidth}
        \centering
        \caption{Comparison of parameter count, training time and Performance(WER) of our architecture with baseline on Librispeech 100h dataset~\cite{librispeech}. Conf.(E=x) represents conformer architecture with x Encoder layers.}
          \label{res:analysis_encoder}
          \centering
         \begin{tabular}{lcccccc}
            \toprule
            \multirow{2}{*}{Method} & \multirow{2}{0.9cm}{ \small{\# of params} } & \multirow{2}{0.8cm}{\small{Train time}} & \multicolumn{2}{c}{ \small{Dev} } & \multicolumn{2}{c}{ \small{Test} } \\
            \cmidrule(r){4-7}
            & & & \small{Clean}  & \small{Other} & \small{Clean} & \small{Other} \\
            \midrule
            \scriptsize{Conf.(E=12)} & \scriptsize{30.6M} & \scriptsize{24h} & \small{7.9} & \small{21.4} & \small{8.4} & \small{22.0} \\
            \midrule
            \scriptsize{Conf.(E=12) + HuBERT-\textsc{Base} + CA} & \scriptsize{31.0M} & \scriptsize{26h} & \small{5.1} & \small{13.0} & \small{5.4} & \small{13.3} \\
            \scriptsize{Conf.(E=8) + HuBERT-\textsc{Base} + CA} & \scriptsize{24.7M} & \scriptsize{22h}  & \textbf{\small{5.0}} & \small{12.9} & \textbf{\small{5.3}} & \small{13.0} \\
            \scriptsize{Conf.(E=4) + HuBERT-\textsc{Base} + CA} & \scriptsize{18.4M} & \scriptsize{17h} & \small{5.4} & \textbf{\small{12.8}} & \small{5.7} & \textbf{\small{12.7}} \\
            \scriptsize{Conf.(E=2) + HuBERT-\textsc{Base} + CA} & \scriptsize{15.2M} & \scriptsize{15h} & \small{5.5} & \textbf{\small{12.7}} & \small{5.5} & \small{12.8} \\
            \bottomrule
      \end{tabular}
    \end{minipage}%
    \hfill
    \begin{minipage}{0.33\textwidth}
        \centering
        \caption{Comparison of Performance(WER) of our approach with baseline on the Dev and Test splits of Tedlium dataset~\cite{tedlium}. We follow the same heuristics from Table~\ref{res:librispeech} to label the experiments.}
      \label{res:tedlium}
      \centering
      \vspace{0.17cm}
      \begin{tabular}{lcc}
        \toprule
            Method & \small{Dev}  & \small{Test} \\
        \midrule

        \scriptsize{Conformer (Conf.)} & \small{10.5} & \small{8.6} \\
        \midrule
        \scriptsize{Conf. + w2v-\textsc{Base} + CA} & \small{9.6} & \small{9.3} \\
        \scriptsize{Conf. + HuBERT-\textsc{Base} + CA} & \small{9.4} & \small{8.9} \\
        \scriptsize{Conf. + HuBERT-\textsc{Large} + CA} & \textbf{\small{7.6}} & \textbf{\small{6.8}} \\
        \bottomrule
      \end{tabular}
    \end{minipage}%
\end{table}

\paragraph{Visualization of Attention:} Figure~\ref{analysis:heatmap} depicts the attention distribution generated by our Cross-Attention(CA) fusion layer for few examples. As can be seen, the layer has learned to predict the alignment between the two representations, indicating that approaches like Convolution or windowed attention that focuses on local attention can also be employed to achieve reasonable improvement. 
 
\paragraph{Performance on unseen dataset:} Table~\ref{res:tedlium} compares the performance between baselines and different variations of our proposed architecture on the Tedlium dataset. We deliberately chose the SSL model that is not exposed to the Tedlium dataset to test the effectiveness of the generated representations on the unseen datasets. Because the \textsc{Large} model is exposed to a much more diverse dataset, we find it to be far more effective than the \textsc{Base} model. Furthermore, we find normalization to be a critical step for out-of-domain datasets.

\section{Conclusion}

In this work, we propose an end-to-end ASR architecture that integrates self-supervised model representations into the speech encoder. We accomplish this with a fusion layer, which can be as simple as framewise addition to a more complex cross-attention mechanism. To evaluate the efficacy of our approach, we conduct experiments on the Librispeech and Tedlium datasets, demonstrating significant reductions in word error rate~(WER) compared to standard baseline models. Furthermore, we perform a thorough analysis of our approach, with a particular focus on the convergence rate, the impact of the number of parameters, and the information captured by the attention mechanism. Through these detailed analyses, we highlight the speed, efficiency, and scalability that our approach achieves in comparison to other baseline methods.

\bibliography{sample}

\end{document}